\documentclass[11pt,a4paper]{article}
\usepackage[hyperref]{emnlp2020}
\usepackage{times}
\usepackage{latexsym}

\usepackage{microtype}

\usepackage{microtype}
\usepackage{bm}
\usepackage{amsmath,amssymb,amsfonts}
\usepackage{enumerate}
\usepackage{algorithm}
\usepackage{algorithmic}
\usepackage{subfigure}
\usepackage{graphicx}
\usepackage{multirow}
\usepackage{booktabs}
\usepackage{arydshln}

\usepackage{xcolor}

\usepackage{cases}

\usepackage{color}
\definecolor{bluei}{RGB}{0,150,255}
\definecolor{redi}{RGB}{183,15,18}
\definecolor{greeni}{RGB}{111,139,62}

\definecolor{mred}{RGB}{255,114,89}
\definecolor{mblue}{RGB}{102, 102, 255}

\aclfinalcopy

\title{Retrofitting Structure-aware Transformer Language Model for End Tasks}

\author{
Hao Fei$^1$, Yafeng Ren$^2$\thanks{~~Corresponding author.} \and Donghong Ji$^1$ \\
1. Department of Key Laboratory of Aerospace Information Security and Trusted Computing,\\
Ministry of Education, School of Cyber Science and Engineering, Wuhan University, China \\ 
2. Guangdong University of Foreign Studies, China \\
\texttt{\{hao.fei,renyafeng,dhji\}@whu.edu.cn}\\
}

\date{}

\begin{document}
\maketitle
\begin{abstract}
We consider retrofitting structure-aware Transformer language model for facilitating end tasks by proposing to exploit syntactic distance to encode both the phrasal constituency and dependency connection into the language model.
A middle-layer structural learning strategy is leveraged for structure integration, accomplished with main semantic task training under multi-task learning scheme.
Experimental results show that the retrofitted structure-aware Transformer language model achieves improved perplexity, meanwhile inducing accurate syntactic phrases.
By performing structure-aware fine-tuning, our model achieves significant improvements for both semantic- and syntactic-dependent tasks.
\end{abstract}

\section{Introduction}

Natural language models (LM) can generate fluent text and encode factual knowledge \cite{MikolovSCCD13,pennington-etal-2014-glove,MerityX0S17}.
Recently, pre-trained contextualized language models have given remarkable improvements on various NLP tasks \cite{PetersNIGCLZ18,GPT2018,howard2018universal,yang2019xlnet,devlin2019bert,dai-etal-2019-transformer}.
Among such methods, the Transformer-based \cite{vaswani2017attention} BERT has become a most popular encoder for obtaining state-of-the-art NLP task performance.
It has been shown \cite{Probing2018,Probing19} that besides rich semantic information, implicit language structure knowledge can be captured by a deep BERT \cite{Analyzing1906,jawahar2019,Assessing190105287}.
However, such structure features learnt via the vanilla Transformer LM are insufficient for those NLP tasks that heavily rely on syntactic or linguistic knowledge \cite{hao-etal-2019-visualizing}.
Some effort devote to improved the ability of structure learning in Transformer LM by installing novel syntax-attention mechanisms \cite{ahmed-etal-2019-need,wang-etal-2019-tree}.
Nevertheless, several limitations can be observed.

\begin{figure}[!t]
\centering
\includegraphics[width=0.75\columnwidth]{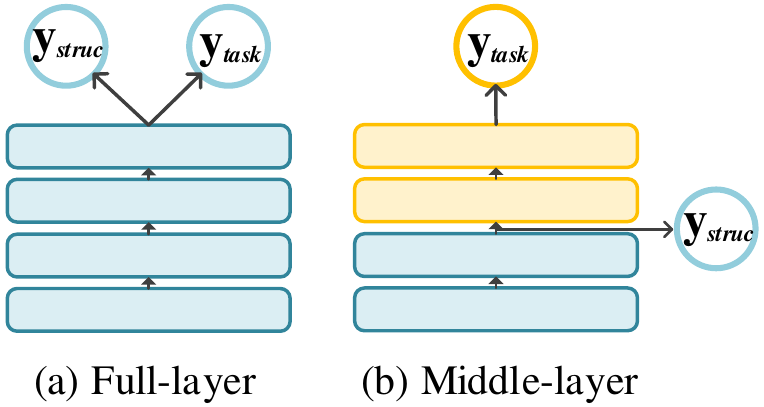}
\caption{
Full-layer multi-task learning for structural training (left), and the middle-layer training for deep structure-aware Transformer LM (right).
}
\label{middle-level training}
\end{figure}

First, according to the recent findings by probing tasks \cite{Probing2018,Probing19,Assessing190105287}, the syntactic structure representations are best retained right at the middle layers \cite{Analyzing1906,jawahar2019}.
Nevertheless, existing tree Transformers employ traditional full-scale training over the whole deep Transformer architecture (as shown in Figure \ref{middle-level training}(a)), consequently weakening the upper-layer semantic learning that can be crucial for end tasks.
Second, these tree Transformer methods encode either  standalone constituency or dependency structure, while different tasks can depend on varying types of structural knowledge.
The constituent and dependency representation for syntactic structure share underlying linguistic characteristics, while the former focuses on disclosing phrasal continuity and the latter aims at indicating dependency relations among elements.
For example, semantic parsing tasks are more dependent on the dependency features \cite{rabinovich-etal-2017-abstract,XiaL0ZFWS19}, while constituency information is much needed for sentiment classification \cite{socher-etal-2013-recursive}.

In this paper, we aim to retrofit structure-aware Transformer LM for facilitating end tasks.
$\bullet$ On the one hand, we propose a structure learning module for Transformer LM,
meanwhile exploiting syntactic distance as the measurement for encoding both the phrasal constituency and the dependency connection.
$\bullet$ On the other hand, as illustrated in Figure \ref{middle-level training}, to better coordinate the structural learning and semantic learning, we employ a middle-layer structural training strategy to integrate syntactic structures to the main language modeling task under multi-task scheme, which encourages the induction of structural information to take place at most suitable layer.
$\bullet$ Last but not least, we consider performing structure-aware fine-tuning with end-task training, allowing learned syntactic knowledge in accordance most with the end task needs.

We conduct experiments on language modeling and a wide range of NLP tasks.
Results show that the structure-aware Transformer retrofitted via our proposed middle-layer training strategy achieves better language perplexity, meanwhile inducing high-quality syntactic phrases.
Besides, the LM after structure-aware fine-tuning can give significantly improved performance for various end tasks, including semantic-dependent and syntactic-dependent tasks.
We also find that supervised structured pre-training brings more benefits to syntactic-dependent tasks, while the unsupervised LM pre-training brings more benefits to semantic-dependent tasks.
Further experimental results on unsupervised structure induction demonstrate that different NLP tasks rely on varying types of structure knowledge as well as distinct granularity of phrases, and our retrofitting method can help to induce structure phrases that are most adapted to the needs of end tasks.

\section{Related Work}

\paragraph{Contextual language modeling.}

Contextual language models pre-trained on a large-scale corpus have witnessed significant advances \cite{PetersNIGCLZ18,GPT2018,howard2018universal,yang2019xlnet,devlin2019bert,dai-etal-2019-transformer}.
In contrast to the traditional static and context-independent word embedding, contextual language models can strengthen word representations by dynamically encoding the contextual sentences for each word during pre-training.
By further fine-tuning with end tasks, the contextualized word representation from language models can help to give the most task-related context-sensitive features \cite{PetersNIGCLZ18}.
In this work, we follow the line of Transformer-based \cite{vaswani2017attention} LM (e.g., BERT), considering its prominence.

\paragraph{Structure induction.}

The idea of introducing tree structures into deep models for structure-aware language modeling has long been explored by supervised structure learning, which generally relies on annotated parse trees during training and maximizes the joint likelihood of sentence-tree pairs \cite{Socher10learningcontinuous,socher-etal-2013-recursive,tai-etal-2015-improved,yazdani-henderson-2015-incremental,dyer-etal-2016-recurrent,Alvarez-MelisJ17,AharoniG17,eriguchi-etal-2017-learning,wang-etal-2018-tree,gu-etal-2018-top}.

There has been much attention paid to unsupervised grammar induction task \cite{parse170901121,ShenLHC18,OrderedShenTSC18,kuncoro-etal-2018-lstms,kim-etal-2019-compound,luo-etal-2019-improving,Unsupervised190402142,kim-etal-2019-unsupervised}.
For example, PRPN \cite{ShenLHC18} computes the syntactic distance of word pairs.
On-LSTM \cite{OrderedShenTSC18} allows hidden neurons to learn long-term or short-term information by a gate mechanism.
URNNG \cite{kim-etal-2019-unsupervised} applies amortized variational inference, encouraging the decoder to generate reasonable tree structures.
DIORA \cite{Unsupervised190402142} uses inside-outside dynamic programming to compose latent representations from all possible binary trees.
PCFG \cite{kim-etal-2019-compound} achieves grammar induction by probabilistic context-free grammar.
Unlike these recurrent network based structure-aware LM, our work focuses on structure learning for a deep Transformer LM.

\paragraph{Structure-aware Transformer language model.}
Some efforts have been paid for the Transformer-based pre-trained language models (e.g. BERT) by visualizing the attention \cite{Analyzing1906,Revealing19,hao-etal-2019-visualizing} or probing tasks \cite{jawahar2019,Assessing190105287}.
They find that the latent language structure knowledge is best retained at the middle-layer in BERT \cite{Analyzing1906,jawahar2019,Assessing190105287}.
\newcite{ahmed-etal-2019-need} employ a decomposable attention mechanism for recursively learn the tree structure for Transformer.
\newcite{wang-etal-2019-tree} integrate tree structures into Transformer via constituency-attention.
However, these Transformer LMs suffer from the full-scale structural training and monotonous types of the structure, limiting the performance of structure LMs for end tasks.
Our work is partially inspired by \newcite{ShenLHC18} and \newcite{luo-etal-2019-improving} on employing syntax distance measurements, while their works focus on the syntax learning by recurrent LMs.

\section{Model}

\begin{figure}[!t]
\centering
\includegraphics[width=0.98\columnwidth]{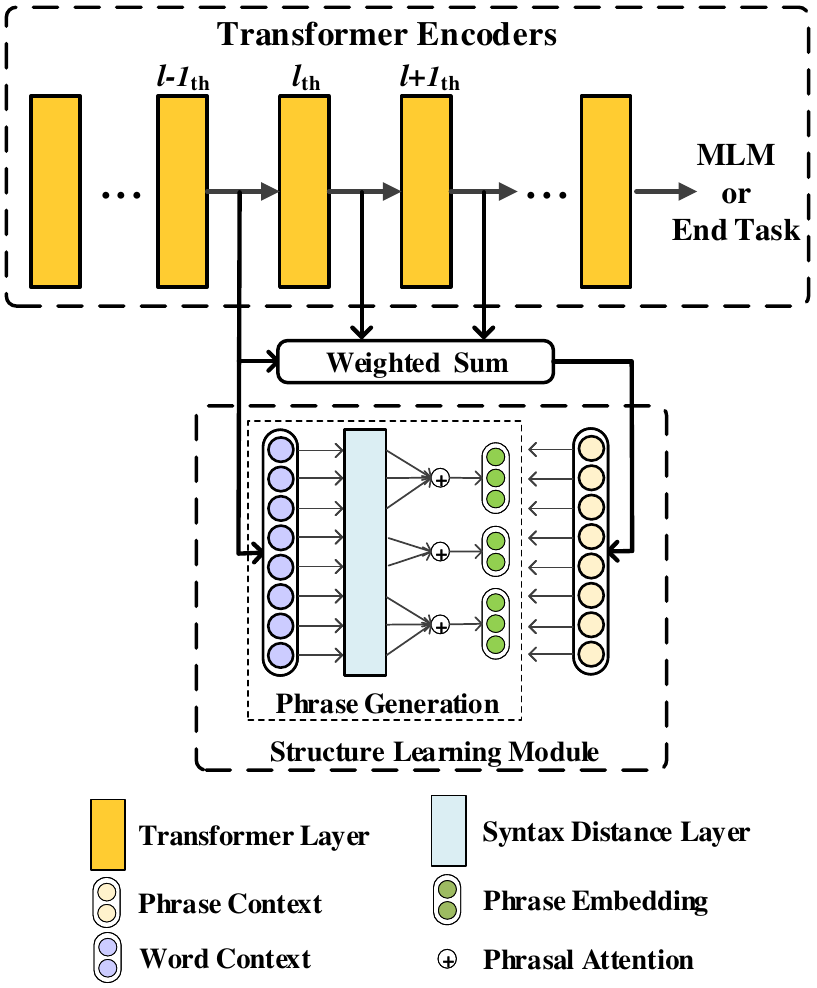}
\caption{
Overall framework of the retrofitted structure-aware Transformer language model.
}
\label{framework}
\end{figure}

The proposed structure-aware Transformer language model mainly consists of two components: the Transformer encoders and structure learning module, which are illustrated in Figure \ref{framework}.

\subsection{Transformer Encoder}

The language model is built based on $N$-layer Transformer blocks.
One Transformer layer applies multi-head self-attention in combination with a feedforward network, layer normalization and residual connections.
Specifically, the attention weights are computed in parallel via:
\begin{equation}
\begin{aligned}
    \bm{E} &= \text{softmax}(\frac{\bm{Q} \bm{K}^{T}}{\sqrt{d}}) \bm{V} \\
             &= \text{softmax}( \frac{(t\cdot\bm{x}) \quad (t\cdot\bm{x})}{\sqrt{d}}) (t\cdot\bm{x})
\end{aligned}
\end{equation}
where $Q$ (query), $K$ (key) and $V$ (value) in multi-head setting process the input $\bm{x} = \{x_{1}, \cdots, x_{n}\}$ $t$ times.

Given an input sentence $\bm{x}$, the output contextual representation of the $l$-th layer Transformer block can be formulated as:
\begin{equation}
\begin{aligned}
    \{\bm{h}^{l}_{1}, \cdots, \bm{h}^{l}_{n}\} &= \text{Trm} (\{x_{1}, \cdots, x_{n}\}) \\
    &= \eta(\Phi(\eta(\bm{E}^{l})) + \bm{E}^{l})
\end{aligned}
\end{equation}
where $\eta$ is the layer normalization operation and $\Phi$ is a feedforward network.
In this work, the output contextual representation $\bm{h}^{l} = \{\bm{h}^{l}_{1}, \cdots, \bm{h}^{l}_{n}\}$ of the middle layers can be used to learn the structure $y_{struc}$, and the one at the final layer will be used for the language modeling or end task training $y_{task}$.

\subsection{Unsupervised Syntax Learning Module}

The structure learning module is responsible for unsupervisedly generating phrases, providing structure-aware language modeling to the host LM.

\paragraph{Syntactic context.}

We extract the context representations from Transformer middle layers for the next syntax learning.
We optimize the structure-aware Transformer LM by forcing the structure knowledge injection focused at middle three layers: $\small{(l-1)_{\text{th}}}$, $\small{l_{\text{th}}}$, and $\small{(l+1)_{\text{th}}}$.
Note that although we only make structural attending to the selected layers, structure learning can enhance lower layers via back-propagation.

Specifically, we take the first of the chosen three-layer as the word context $\bm{C}^{\Psi} = \bm{h}^{l-1}$.
For the phrasal context $\bm{C}^{\Omega} = \{\bm{c}^{\Omega}_{1}, \cdots, \bm{c}^{\Omega}_{n}\}$, we make use of contextual representations from the three chosen layers by weighted sum:
\begin{equation}
    \bm{C}^{\Omega} = \alpha_{l-1}\cdot\bm{h}^{l-1} + \alpha_{l}\cdot\bm{h}^{l} + \alpha_{l+1}\cdot\bm{h}^{l+1}
\end{equation}
where $\alpha_{l-1}$, $\alpha_{l}$ and $\alpha_{l+1}$ are sum-to-one trainable coefficients.
Rich syntactic representations are expected to be captured in $\bm{C}^{\Omega}$ by LM.

\paragraph{Structure measuring.}

\begin{figure}[!t]
\centering
\includegraphics[width=0.96\columnwidth]{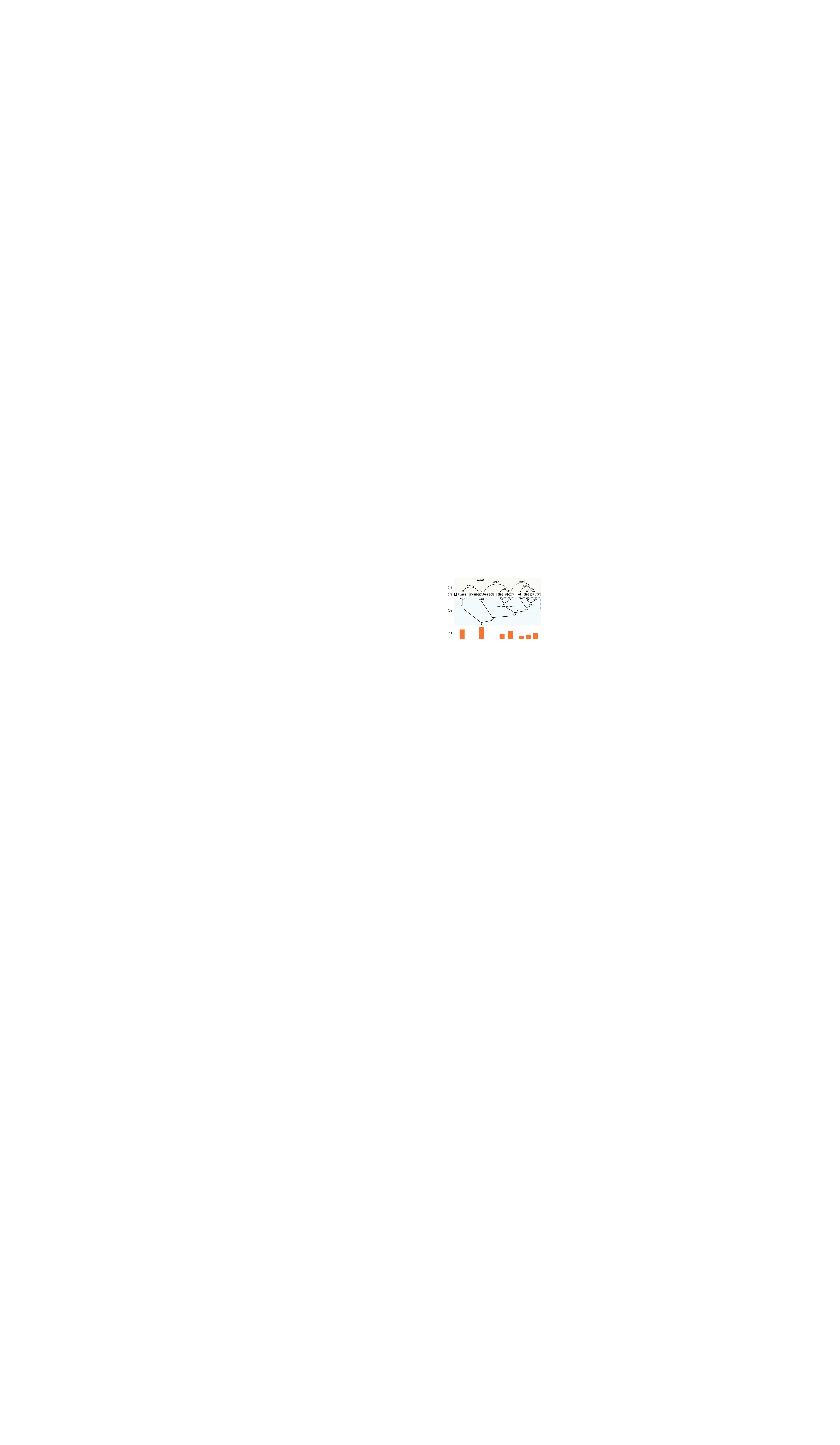}
\caption{
Simultaneously measuring dependency relations (1) and phrasal constituency (3) based on the example sentence (2) by employing syntax distance (4).
}
\label{syntax-distance}
\end{figure}

In this study, we reach the goal of measuring syntax by employing \emph{\textbf{syntax distance}}.
The general concept of syntax distance $d_i$ can be reckoned as a metric (i.e., distance) from a certain word $x_i$ to the root node within the dependency tree \cite{ShenLHC18}.
For instance in Figure \ref{syntax-distance}, the head word `\emph{remembered}' $x_{i}$ and its dependent word `\emph{James}'  $x_{j}$ follow $d_{i} < d_{j}$.
While in this work, to maintain both the dependency and phrasal constituents simultaneously, we add additional constraints on words and phrases.
Given two words $x_{i}$ and $x_{j}$ ($\small{0\le i<j\le n}$) in one phrase, we define $d_{i} < d_{j}$.
This can be demonstrated by the word pair `\emph{the}' and `\emph{story}'.
While if they are in different phrases\footnote{Note that we cannot explicitly define the granularity (width) of every phrases in constituency tree, while instead it will be decided by the structure learning module in heuristics. }, e.g., $S_u$ and $S_v$, the corresponding inner-phrasal head words follow $d_{i}$ (in $S_u$) $>$ $d_{j}$ (in $S_V$), e.g., `\emph{story}' and `\emph{party}'.

In the structure learning module, we first compute the syntactic distances $\bm{d} = \{d_{1}, \cdots, d_{n}\}$ for each word based on the word context via a convolutional network:
\begin{equation}
\label{syntax-dis}    \{d_{1}, \cdots, d_{n}\} = \Phi(\text{CNN}(\{\bm{c}^{\Psi}_{1}, \cdots, \bm{c}^{\Psi}_{n}\}))
\end{equation}
where $d_{i}$ is a scalar, and $\Phi$ is for linearization.
With such syntactic distance, we expect both the dependency as well as constituency syntax can be well captured in LM.

\paragraph{Syntactic phrase generating.}

Considering the word $x_i$ opening an induced phrase $S_{m} = [x_{i}, \cdots, x_{i+w}]$ in a sentence, where $w$ is the phrase width, we need to decide the probability $p^{*}(x_j)$ that a word $x_{j}$ ($\small{j}$=$\small{i+w+1}$) (i.e., the first word outside phrase $S_{m}$) belongs to $S_{m}$:
\begin{equation}
    p^{*}(x_j) =  \prod_{k=i}^{i+w}  \text{sigmoid}(d_j - d_k).
\end{equation}
We set the initial width $w=1$, if $p^{*}(x_{j})$ is above the window threshold $\lambda$, $x_{j}$ should be considered inside the phrase; otherwise, the phrase $S_{m}$ should be closed and restart at $x_{j}$.
We incrementally conduct such phrasal searching procedure to segment all the phrases in a sentence.
Given an induced phrase $S_{m} = [x_{i}, \cdots, x_{i+w}]$, we obtain its embedding $\bm{s}_{m}$ via a phrasal attention:
\begin{align}
\label{u_i}    u_i &= \text{softmax}(d_i \cdot p^{*}(x_i)) \\
\label{s_m} \bm{s}_m &=  \sum_{i}^{i+w}   u_i \cdot \bm{c}^{\Psi}_i
\end{align}

\section{Structure-aware Learning}

\paragraph{Multi-task training for language modeling and structure induction.}
Different from traditional language models, a Transformer-based LM employs the masked language modeling (MLM), which can capture larger contexts.
Likewise, we predict a masked word using the corresponding context representation at the top layer:
\begin{align}
    p^{\text{W}}(y_i| \bm{x}) &=  \text{softmax}(\bm{c_i} | \bm{x})  \\
   \mathcal{L}_{\text{W}} &=  \sum_{i}^{k}  \log p^{\text{W}}(y_i| \bm{x})
\end{align}

On the other hand, the purpose on unsupervised syntactic induction is to encourage the model to induce $\bm{s}_m$ that is most likely entailed by the phrasal context $\bm{c}^{\Omega}_i$.
The behind logic lies is that, if the initial Transformer LM can capture linguistic syntax knowledge, then after iterations of learning with the structure learning module, the induced structure can be greatly amplified and enhanced \cite{luo-etal-2019-improving}.
We thus define the following probability:
\begin{equation}
    p^{\text{G}}(\bm{s}_m|\bm{c}^{\Omega}_i) = \frac{1}{ 1 + \exp(- \bm{s}^{T}_m \cdot \bm{c}^{\Omega}_i )}
\end{equation}

Additionally, to enhance the syntax learning, we employ negative sampling:
\begin{equation}
    \mathcal{L}_{\text{Neg}} = \frac{1}{n}  \sum_j^n   p^{\text{G}}(\bm{\hat{s}}^{T}_j|\bm{c}^{\Omega}_i)
\end{equation}
where $\hat{s}$ is a randomly selected negative phrase.
The final objective for structure learning is:
\begin{equation}
 \mathcal{L}_{\text{G}} = \sum_i^{K}( \sum^{\text{M}}_{m} (1- p^{\text{G}}(\bm{s}_m|\bm{c}^{\Omega}_i) ) + \mathcal{L}_{\text{Neg}} )
\end{equation}

We employ multi-task learning for simultaneously training our LM for both word prediction and structure induction.
Thus, the overall target is to minimize the following multi-task loss objective:
\begin{equation}
  \mathcal{L}_{\text{pre}} =  \mathcal{L}_{\text{W}} + \gamma^{\text{pre}} \cdot \mathcal{L}_{\text{G}}
\end{equation}
where $\gamma^{\text{pre}}$ is a regulating coefficient.

\paragraph{Supervised syntax injection.}
Our default structure-aware LM unsupervisedly induces syntax at the pre-training stage, as elaborated above.
Alternatively, in Eq. (\ref{s_m}), if we leverage the gold (or apriori) syntax distance information for phrases, we can achieve supervised structure injection.

\paragraph{Unsupervised structure fine-tuning.}

We aim to improve the learnt structural information for better facilitating the end tasks.
Therefore, during the fine-tuning stage of end tasks, we consider further making the structure learning module trainable:
\begin{equation}
  \mathcal{L}_{\text{fine}} =  \mathcal{L}_{\text{task}} + \gamma^{\text{fine}} \cdot \mathcal{L}_{\text{G}}
\end{equation}
where $\mathcal{L}_{\text{task}}$ refers to the loss function of the end task, and $\gamma^{\text{fine}}$ is a regulating coefficient.
Note that to achieve the best structural fine-tuning, the supervised structure injection is unnecessary, and we do not allow supervised structure aggregation at the fine-tuning stage.

Our approach is model-agnostic as we realize the syntax induction via a standalone structure learning module, which is disentangled from a host LM.
Thus the method can be applied to various Transformer-based LM architectures.

\section{Experiments}

\subsection{Experimental Setups}

We employ the same architecture as BERT base model\footnote{\url{https://github.com/google-research/bert}}, which is a 12-layer Transformer with 12 attention heads and 768 dimensional hidden size.
To enrich our experiments, we also consider the Google pre-trained weights as the initialization.
We use Adam as our optimizer with an initial learning rate in [8e-6, 1e-5, 2e-5, 3e-5], and a L2 weight decay of 0.01.
The batch size is selected in [16,24,32].
We set the initial values of coefficients $\alpha_{l-1}, \alpha_{l}$ and $\alpha_{l+1}$ as 0.35, 0.4 and 0.25, respectively.
The pre-training coefficient $\gamma^{\text{pre}}$ is set as 0.5, and the fine-tuning one $\gamma^{\text{fine}}$ as 0.23.
These values give the best effects in our development experiments.
Our implementation is based on the PyTroch library\footnote{\url{https://pytorch.org/}}.

Besides, for supervised structure learning in our experiments, we use the state-of-the-art BiAffine dependency parser \cite{BiaffineDozatM17} to parse sentences for all the relevant datasets, and use the Self-Attentive parser \cite{kitaev-klein-2018-constituency} to obtain the constituency structure.
Being trained on the English Penn Treebank (PTB) corpus \cite{PTB93}, the dependency parser has 95.2\% UAS and 93.4\% LAS, and the constituency parser has 92.6\% F1 score.
With the auto-parsed annotations, we can calculate the syntax distances (substitute the ones in Eq. \ref{syntax-dis}) and obtain the corresponding phrasal embeddings (in Eq. \ref{s_m}).

\begin{figure}[!t]
\centering
\includegraphics[width=1.0\columnwidth]{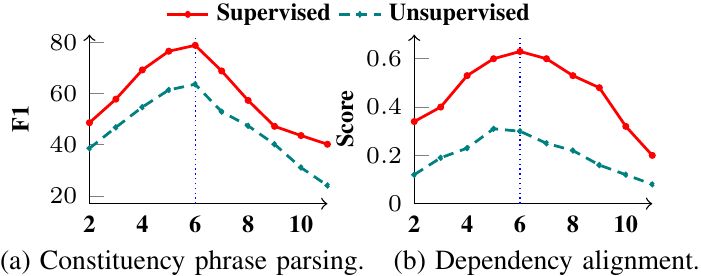}
\caption{
Development experiments on syntactic probing tasks at varying Transformer layer.
}
\label{probing}
\end{figure}
\subsection{Development Experiments}

\paragraph{Structural learning layers.}

We first validate at which layer of depths the structural-aware Transformer LM can achieve the best performance when integrating our retrofitting method.
We thus design probing experiments, in which we consider following two syntactic tasks.
1) \textbf{Constituency phrase parsing} seeks to generate grammar phrases based on the PTB dataset and evaluate whether induced constituent spans also exist in the gold Treebank dataset.
2) \textbf{Dependency alignment} aims to compute the proportion of Transformer attention connecting tokens in a dependency relation \cite{Analyzing1906}:
\begin{equation}
    \text{Score} = \frac{
    \sum_{x \in X}\sum_{i=1}^x \sum_{j=1}^x \alpha_{i,j}(x) \cdot \text{dep}(x_i, x_j)
    }{
    \sum_{x \in X}\sum_{i=1}^x \sum_{j=1}^x \alpha_{i,j}(x)
    }
\end{equation}
where $\small{\alpha_{i,j}(x)}$ is the attention weight, and $\text{dep}(x_i, x_j)$ is an indicator function (1 if $x_i$ and $x_j$ are in a dependency relation and 0 otherwise).
The experiments are based on English Wikipedia, following Vig and Belinkov \shortcite{Analyzing1906}.

\begin{figure}[!t]
\centering \includegraphics[width=.89\columnwidth]{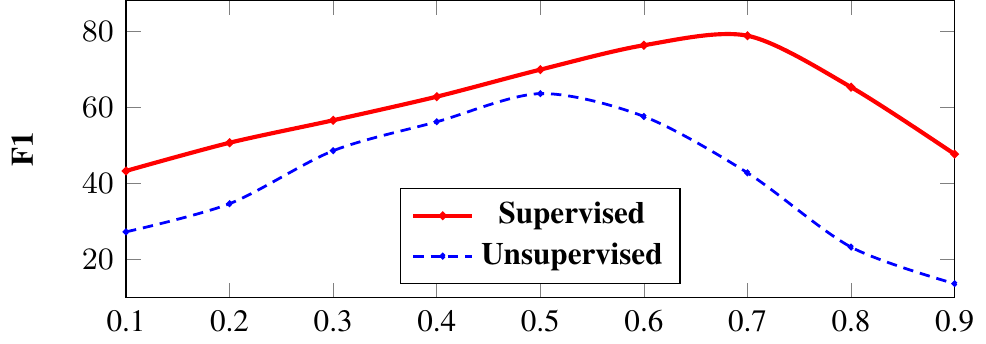}
\caption{
Constituency parsing under different $\lambda$.
}
\label{lambda}
\end{figure}

As shown in Figure \ref{probing}, both the results on unsupervised and supervised phrase parsing are the best at layer 6.
Also the attention aligns with dependency relations most strongly in the middle layers (5-6), consistent with findings from previous work \cite{Probing19,Analyzing1906}.
Both two probing tasks indicate that our proposed middle-layer structure training is practical.
We thus inject the structure in the structure learning module at the $6$-th layer ($l=6$).

\begin{table*}[!t]
\begin{center}
\resizebox{1.86\columnwidth}{!}{
  \begin{tabular}{lccccccccc}
\hline
\multirow{2}{*}{ system } & \multicolumn{3}{c}{ \texttt{Syntactic.}}&   \multicolumn{5}{c}{ \texttt{Semantic.}} & \multirow{2}{*}{ \emph{Avg.}}\\
\cmidrule(r){2-4}\cmidrule(r){5-9} 
& TreeDepth &   TopConst &    Tense &   SOMO &    NER  &  SST &   Rel &   SRL &\\
\hline
\multicolumn{9}{l}{ $\bullet$ \textbf{w/o Initial Weight:}}\\
Trm &   25.31 &   40.32 &   61.06 &   50.11 &   89.22 &   86.21 &   84.70 &   88.30 &65.65\\
RvTrm &    29.52 &   45.01 &   63.83 &   51.42 &   89.98 &   86.66 &   85.02 &   88.94 &67.55\\
Tree+Trm &    30.37 &   46.58 &   65.83 &   53.08 &   90.62 &   87.25 &   84.97 &   88.70 &  68.43\\
PI+TrmXL &    31.28 &   47.06 &   63.78 &   52.36 &   90.34 &   87.09 &   85.22 &   89.02 &  68.27\\
\hdashline
\multicolumn{9}{l}{Ours+Trm }\\
\quad +\textbf{usp.} &    33.98 &   49.69 &   66.39 &   \underline{57.04} &    \underline{92.24} &   \underline{90.48}&   \underline{87.05} &   \underline{90.87} &  70.74\\
\quad +\textbf{sp.} &   \underline{37.35} &   \underline{57.68} &   \underline{72.04} &   {56.41} &    91.86&     90.06&  86.34  &   90.54 &  73.12\\
\quad +\textbf{syn-embed.}  &    36.28 &   54.30 &   67.61 &   55.68 &   91.87 &   87.10 &   86.87 &   89.41 &  71.14\\
\hline\hline
\multicolumn{9}{l}{ $\bullet$ \textbf{Initial Weight:}}\\
BERT&38.61 &    79.37 &   90.61 &   65.31 &   92.40 &   93.50 &   89.25 &   92.20  &  80.16\\
Ours+BERT(\textbf{usp.})& \underline{45.82} &    \underline{88.64} &   \underline{94.68} &   \underline{67.84} &   \underline{94.28} &   \underline{94.67} &   \underline{90.41} &   \underline{93.12}  &  83.68\\
\hline
\end{tabular}
}
\end{center}
  \caption{
  Structure-aware Transformer LM for end tasks.
  }
  \label{fine-tune end task}
\end{table*}

\begin{table}[!t]
\begin{center}
\resizebox{0.65\columnwidth}{!}{
  \begin{tabular}{lcc}
\hline
System &  Const. & Ppl. \\
\hline\hline
PRPN& 42.8& - \\
On-LSTM&  49.4& - \\
URNNG&  52.4& -  \\
DIORA&  56.2& - \\
PCFG& 60.1& - \\
\hline
Trm&  22.7& 78.6 \\
RvTrm & 47.0& 50.3 \\
Tree+Trm & 52.0& 45.7 \\
PI+TrmXL & 56.2& 43.4 \\
\cdashline{1-3}[0.8pt/2pt]
\multicolumn{3}{l}{Ours+Trm}\\
\quad +\textbf{usp.}& 60.3& 37.0 \\
\quad +\textbf{sp.}&  68.8& 29.2 \\
\hline\hline
BERT& 31.3& 21.5 \\
Ours+BERT(\textbf{usp.})& 65.2& 16.2 \\
\hline
\end{tabular}
}
\end{center}
\caption{
Performance on constituency parsing and language modeling.
}
\label{Structure-aware Language Modeling}
\end{table}

\paragraph{Phrase generation threshold.}

We introduce a hyper-parameter $\lambda$ as a threshold to decide whether a word belong to a given phrase during the phrasal generation step.
We explore the best $\lambda$ value based on the same parsing tasks.
As shown in Figure \ref{lambda}, with $\lambda=0.5$ for unsupervised induction and $\lambda=0.7$ for supervised induction, the induced phrasal quality is the highest.
Therefore we set such $\lambda$ values for all the remaining experiments.

\subsection{Structure-aware Language Modeling}

We evaluate the effectiveness of our proposed retrofitted structure-aware LM after pre-training.
We first compare the performance on language modeling\footnote{Transformer can see its subsequent words bidirectionally, so we measure the perplexity on masked words.
And we thus avoid directly comparing with the Recurrent-based LMs.}.
From the results shown in Table \ref{Structure-aware Language Modeling}, our retrofitted Transformer yields better language perplexity in both unsupervised (37.0) or supervised (29.2) manner.
This proves that our middle-layer structure training strategy can effectively relieve negative mutual influence of structure learning on semantic learning, while inducing high-quality of structural phrases.
We can also conclude that language models with more successful structural knowledge can better help to encode effective intrinsic language patterns, which is consistent with the prior studies \cite{kim-etal-2019-unsupervised,wang-etal-2019-tree,Unsupervised190402142}.

We also compare the constituency parsing with state-of-the-art structure-aware models, including \textbf{1) Recurrent-based models} described in $\S$2: PRPN \cite{ShenLHC18}, On-LSTM \cite{OrderedShenTSC18}, URNNG \cite{kim-etal-2019-unsupervised}, DIORA \cite{Unsupervised190402142}, PCFG \cite{kim-etal-2019-compound}, and \textbf{2) Transformer based methods}: Tree+Trm \cite{wang-etal-2019-tree}, RvTrm \cite{ahmed-etal-2019-need},
PI+TrmXL \cite{luo-etal-2019-improving}, and the BERT model initialized with rich weights.
As shown in Table \ref{Structure-aware Language Modeling}, all the structure-aware models can give good parsing results, compared with non-structured models.
Our retrofitted Transformer LM gives the best performance (60.3\% F1) in unsupervised induction.
Combined with the supervised auto-labeled parses, it give the highest F1 score (68.8\%).

\begin{figure*}[!t]
\centering
\includegraphics[width=2.1\columnwidth]{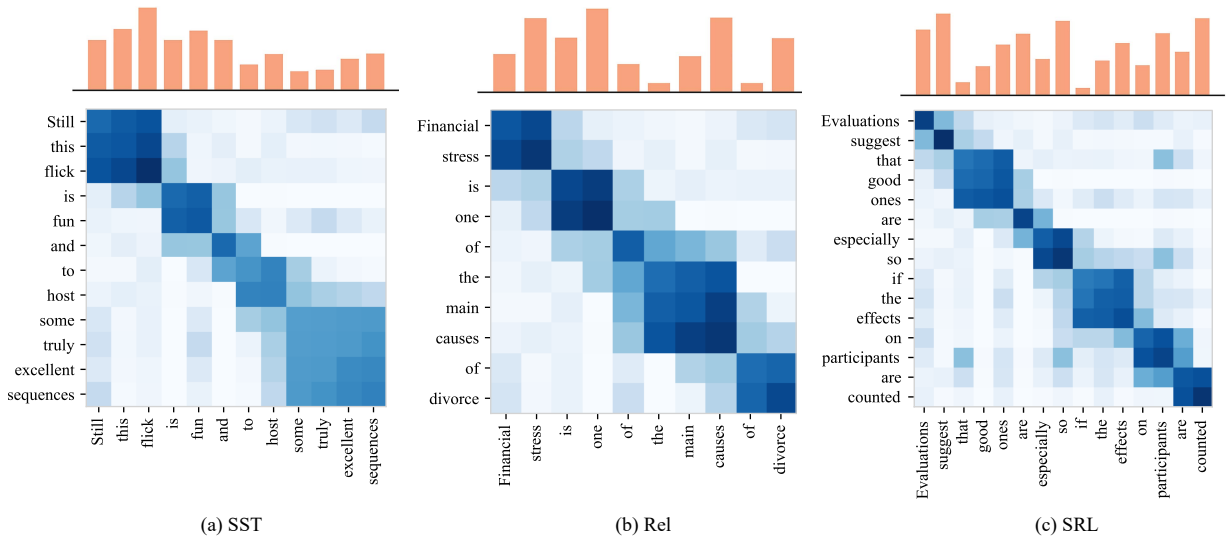}
\caption{
Visualization of attention heads (heatmap) and the corresponding syntax distances (bar chart).
}
\label{visualization}
\end{figure*}

\subsection{Fine-tuning for End Tasks}

We validate the effectiveness of our method for end tasks with structure-aware fine-tuning.
All systems are first pre-trained for structure learning, and then fine-tuned with end task training.
The evaluation is performed on eight tasks, involving syntactic tasks and semantic tasks.
\texttt{TreeDepth} predicts the depth of the syntactic tree, \texttt{TopConst} tests the sequence of top level constituents in the syntax tree, and \texttt{Tense} detects the tense of the main-clause verb, while \texttt{SOMO} checks the sensitivity to random replacement of words, which are the standard probing tasks.
We follow the same datasets and settings with previous work \cite{Probing2018,jawahar2019}.

Also we evaluate the semantic tasks including 1) \texttt{NER}, named entity recognition on CoNLL03 \cite{CoNLL2003}, 2) \texttt{SST}, binary sentiment classification task on Standford sentiment treebank \cite{socher-etal-2013-recursive}, 3) \texttt{Rel}, relation classification on Semeval10 \cite{hendrickx2009semeval}, and 4) \texttt{SRL}, semantic role labeling task on the CoNLL09 WSJ \cite{conll2009}.
The performance is reported by the F1 score.

The results are summarized in Table \ref{fine-tune end task}.
First, we find that structure-aware LMs bring improved performance for all the tasks, compared with the vanilla Transformer encoder.
Second, the Transformer with our structural-aware fine-tuning achieves better results (70.74\% on average) for all the end tasks, compared with the baseline tree Transformer LMs.
This proves that our proposed middle-layer strategy best benefits the structural fine-tuning, compared with the full-layer structure training on baselines.
Third, with supervised structure learning, significant improvements can be found across all tasks.

For the supervised setting, we replace the supervised syntax fusion in structure learning module with the auto-labeled syntactic dependency embedding and concatenate it with other input embeddings.
The results are not as prominent as the supervised syntax fusion, which reflects the advantage of our proposed structure learning module.
Besides, based on the task improvements from the retrofitted Transformer by our method, we can further infer that the supervised structure benefits more syntactic-dependent tasks, and the unsupervised structure benefits semantic-dependent tasks the most.
Finally, the BERT model integrating with our method can give improved effects\footnote{We note that the direct comparison with BERT model is not fair, because the large numbers of well pre-trained parameters can bring overwhelming advances.}.

\begin{table}[!t]
\begin{center}
\resizebox{0.64\columnwidth}{!}{
\begin{tabular}{lcc}
\hline
& Mean& Median \\
\hline
RvTrm& 0.68& 0.69 \\
Tree+Trm& 0.60& 0.64 \\
PI+TrmXL& 0.54& 0.58 \\
Ours+Trm(\textbf{usp.}) & 0.50& 0.52 \\
\cdashline{1-3}[0.8pt/2pt]
Ours+Trm(\textbf{sp.})  & 0.32& 0.37 \\
\hline
\end{tabular}
}
\end{center}
  \caption{Fine-grained parsing. }
  \label{parsing fine-grained}
\end{table}

\section{Analysis}

\subsection{Induced Phrase after Pre-training.}
We take a further step, evaluating the fine-grained quality on phrasal structure induction after pre-training.
Instead of checking whether the induced constituent spans are identical to the gold counterparts, we now consider measuring the deviation ${PhrDev}(\hat{y}, y) = \sqrt{\frac{1}{N} \sum_i [\Delta(\hat{y}_i, y_i) - \overline{\Delta} ]^{2} }$,
where $\Delta(\hat{y}_i, y_i)$ is the phrasal editing distance between the induced phrase length and the gold length within a sentence.
$\overline{\Delta}$ is the averaged editing distance.
If all the predicted phrases are same with the ground truth, or all different from it, ${PhrDev}(\hat{y}, y)=0$, which means that the phrases are induced with the maximum consistency, and vice versa.
We make statistics for all the sentences in Table \ref{parsing fine-grained}.
Our method can unsupervisedly generate higher quality of structural phrases, while we can achieve the best injection of the constituency knowledge into LM by the supervised manner.

\subsection{Fine-tuned Structures with End Tasks}

\paragraph{Interpreting fine-tuned syntax.}

To interpret the fine-tuned structures, we empirically visualize the Transformer attention head from the chosen $l$-layer, and the syntax distances of the sentence.
We exhibit three examples from \texttt{SST}, \texttt{Rel} and \texttt{SRL}, respectively, as shown in Figure \ref{visualization}.
Overall, our method can help to induce clear structure of both dependency and constituency.
While interestingly, different types of tasks rely on different granularity of phrase.
Comparing the heat maps and syntax distances with each other, the induced phrasal constituency on \texttt{SST} are longer than that on \texttt{SRL}.
This is because the sentiment classification task demands more phrasal composition features, while the SRL task requires more fine-grained phrases.
In addition, we find that the syntax distances in \texttt{SRL} and \texttt{Rel} are higher in variance,
compared with the ones on \texttt{SST},
Intuitively, the larger deviation of syntax distances in a sentence indicates the more demand to the interdependent information between elements, while the smaller deviation refers to phrasal constituency.
This reveals that \texttt{SRL} and \texttt{Rel} rely more on the dependency syntax, while \texttt{SST} is more relevant to constituents, which is consistent with previous studies \cite{socher-etal-2013-recursive,rabinovich-etal-2017-abstract,XiaL0ZFWS19,Crossfei9165903}.

\begin{figure}[!t]
\centering \includegraphics[width=.94\columnwidth]{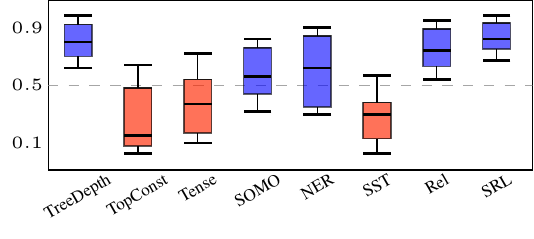}
\caption{
Distributions of dependency and constituency syntax in different tasks.
\textcolor{mblue}{Blue} color indicates the predominance of dependency, while \textcolor{mred}{Red} for constituency.
}
\label{distribution}
\end{figure}

\paragraph{Distributions of heterogeneous syntax for different tasks.}

Based on the above analysis, we further analyze the distributions of dependency and constituency structures after fine-tuning, in different tasks.
Technically, we calculate the mean absolute differences of syntax distances between elements $x_i$ and the sub-root node $x_r$ in a sentence: $Diff=\frac{1}{N}\sum_i^N |d_i - d_r|$.
We then linearly normalize them into [0,1] for all the sentences in the corpus of each task, and make statistics, as plotted in Figure \ref{distribution}.
Intuitively, the larger the value is, the more interdependent to dependency syntax the task is, and otherwise, to constituency structure.
Overall, distributions of dependency structures and phrasal constituents in fine-tuned LM vary among different tasks, verifying that different tasks depend on distinct types of structural knowledge.
For example, \texttt{TreeDepth}, \texttt{Rel} and \texttt{SRL} are most supported by dependency structure, while \texttt{TopConst} and \texttt{SST} benefit from constituency the most.
\texttt{SOMO} and \texttt{NER} can gain from both two types.

\begin{table}[!t]
\begin{center}
\resizebox{1.0\columnwidth}{!}{
\begin{tabular}{lcccc}
\hline
\multirow{2}{*}{  } & \multicolumn{2}{c}{ \texttt{SST}}&   \multicolumn{2}{c}{\texttt{SRL}} \\
\cmidrule(r){2-3}\cmidrule(r){4-5}
& Ours+Trm &  Tree+Trm&    Ours+Trm &   Tree+Trm\\
\hline
\emph{NP} &   0.48&   0.45&   0.37&   0.53 \\
\emph{VP}&   0.21&   0.28&   0.36&   0.21 \\
\emph{PP}&   0.08&   0.14&   0.17&   0.06 \\
\emph{ADJP}&   0.10&   0.05&   0.05&   0.12 \\
\emph{ADVP}&   0.07&   0.02&   0.03&   0.02 \\
Other&  0.06&   0.06&   0.02&   0.06 \\
\hline
Avg.Len.&   3.88&   3.22&   2.69&   3.36\\
\hline
\end{tabular}
}
\end{center}
  \caption{
  Proportion of each type of induced phrase.
  }
  \label{Proportion}
\end{table}

\paragraph{Phrase types.}

Finally, we explore the diversity of phrasal syntax required by two representative end tasks, \texttt{SST} and \texttt{SRL}.
We first look into the statistical proportion for different types of induced phrases\footnote{Five main types are considered: noun phrase (\emph{NP}), verb phrase (\emph{VP}), prepositional phrase (\emph{PP}), adjective phrase (\emph{ADJP}) and adverb phrase (\emph{ADVP}).}.
As shown in Table \ref{Proportion}, our method tends to induce more task-relevant phrases, where the lengths of induced phrases are more variable to the task.
Concretely, the fine-tuned structure-aware Transformer helps to generate more \emph{NP} also with longer phrases for the \texttt{SST} task, and yield roughly equal numbers of \emph{NP} and \emph{VP} for \texttt{SRL} tasks with shorter phrases.
This evidently gives rise to the better task performance.
In contrast, the syntax phrases induced by the Tree+Trm model keep unvarying for \texttt{SST} (3.22) and \texttt{SRL} (3.36) tasks.

\section{Conclusion}

We presented a retrofitting method for structure-aware Transformer-based language model.
We adopted the syntax distance to encode both the constituency and dependency structure.
To relieve the conflict of structure learning and semantic learning in Transformer LM, we proposed a middle-layer structure learning strategy under a multi-tasks scheme.
Results showed that structure-aware Transformer retrofitted via our proposed method achieved better language perplexity, inducing high-quality syntactic phrase.
Furthermore, our LM after structure-aware fine-tuning gave significantly improved performance for both semantic-dependent and syntactic-dependent tasks, also yielding most task-related and interpretable syntactic structures.

\section{Acknowledgments}

We thank the anonymous reviewers for their valuable and detailed comments.
This work is supported by the National Natural Science Foundation of China (No. 61772378, No. 61702121), 
the National Key Research and Development Program of China (No. 2017YFC1200500), 
the Research Foundation of Ministry of Education of China (No. 18JZD015), 
the Major Projects of the National Social Science Foundation of China (No. 11\&ZD189),
the Key Project of State Language Commission of China (No. ZDI135-112) 
and Guangdong Basic and Applied Basic Research Foundation of China (No. 2020A151501705).

\bibliography{emnlp2020}

\begin{thebibliography}{46}
\expandafter\ifx\csname natexlab\endcsname\relax\def\natexlab#1{#1}\fi

\bibitem[{Aharoni and Goldberg(2017)}]{AharoniG17}
Roee Aharoni and Yoav Goldberg. 2017.
\newblock Towards string-to-tree neural machine translation.
\newblock \emph{CoRR}, abs/1704.04743.

\bibitem[{Ahmed et~al.(2019)Ahmed, Samee, and Mercer}]{ahmed-etal-2019-need}
Mahtab Ahmed, Muhammad~Rifayat Samee, and Robert~E. Mercer. 2019.
\newblock You only need attention to traverse trees.
\newblock In \emph{Proceedings of the ACL}, pages 316--322.

\bibitem[{Alvarez{-}Melis and Jaakkola(2017)}]{Alvarez-MelisJ17}
David Alvarez{-}Melis and Tommi~S. Jaakkola. 2017.
\newblock Tree-structured decoding with doubly-recurrent neural networks.
\newblock In \emph{Proceedings of the ICLR}.

\bibitem[{Conneau et~al.(2018)Conneau, Kruszewski, Lample, Barrault, and
  Baroni}]{Probing2018}
Alexis Conneau, German Kruszewski, Guillaume Lample, Lo{\"\i}c Barrault, and
  Marco Baroni. 2018.
\newblock What you can cram into a single {\$}{\&}!{\#}* vector: Probing
  sentence embeddings for linguistic properties.
\newblock In \emph{Proceedings of the ACL}, pages 2126--2136.

\bibitem[{Dai et~al.(2019)Dai, Yang, Yang, Carbonell, Le, and
  Salakhutdinov}]{dai-etal-2019-transformer}
Zihang Dai, Zhilin Yang, Yiming Yang, Jaime Carbonell, Quoc Le, and Ruslan
  Salakhutdinov. 2019.
\newblock Transformer-{XL}: Attentive language models beyond a fixed-length
  context.
\newblock In \emph{Proceedings of the ACL}, pages 2978--2988.

\bibitem[{Devlin et~al.(2019)Devlin, Chang, Lee, and
  Toutanova}]{devlin2019bert}
Jacob Devlin, Ming-Wei Chang, Kenton Lee, and Kristina Toutanova. 2019.
\newblock {BERT}: Pre-training of deep bidirectional transformers for language
  understanding.
\newblock In \emph{Proceedings of the NAACL}, pages 4171--4186.

\bibitem[{Dozat and Manning(2017)}]{BiaffineDozatM17}
Timothy Dozat and Christopher~D. Manning. 2017.
\newblock Deep biaffine attention for neural dependency parsing.
\newblock In \emph{Proceedings of the ICLR}.

\bibitem[{Drozdov et~al.(2019)Drozdov, Verga, Yadav, Iyyer, and
  McCallum}]{Unsupervised190402142}
Andrew Drozdov, Patrick Verga, Mohit Yadav, Mohit Iyyer, and Andrew McCallum.
  2019.
\newblock Unsupervised latent tree induction with deep inside-outside recursive
  autoencoders.
\newblock \emph{CoRR}, abs/1904.02142.

\bibitem[{Dyer et~al.(2016)Dyer, Kuncoro, Ballesteros, and
  Smith}]{dyer-etal-2016-recurrent}
Chris Dyer, Adhiguna Kuncoro, Miguel Ballesteros, and Noah~A. Smith. 2016.
\newblock Recurrent neural network grammars.
\newblock In \emph{Proceedings of the NAACL}, pages 199--209.

\bibitem[{Eriguchi et~al.(2017)Eriguchi, Tsuruoka, and
  Cho}]{eriguchi-etal-2017-learning}
Akiko Eriguchi, Yoshimasa Tsuruoka, and Kyunghyun Cho. 2017.
\newblock Learning to parse and translate improves neural machine translation.
\newblock In \emph{Proceedings of the ACL}, pages 72--78.

\bibitem[{Fei et~al.(2020)Fei, Zhang, Li, and Ji}]{Crossfei9165903}
Hao Fei, Meishan Zhang, Fei Li, and Donghong Ji. 2020.
\newblock Cross-lingual semantic role labeling with model transfer.
\newblock \emph{IEEE/ACM Transactions on Audio, Speech, and Language
  Processing}, 28:2427--2437.

\bibitem[{Goldberg(2019)}]{Assessing190105287}
Yoav Goldberg. 2019.
\newblock Assessing bert's syntactic abilities.
\newblock \emph{CoRR}, abs/1901.05287.

\bibitem[{G{\=u} et~al.(2018)G{\=u}, Shavarani, and Sarkar}]{gu-etal-2018-top}
Jetic G{\=u}, Hassan~S. Shavarani, and Anoop Sarkar. 2018.
\newblock Top-down tree structured decoding with syntactic connections for
  neural machine translation and parsing.
\newblock In \emph{Proceedings of the EMNLP}, pages 401--413.

\bibitem[{Haji{\v{c}} et~al.(2009)Haji{\v{c}}, Ciaramita, Johansson, Kawahara,
  Mart{\'\i}, M{\`a}rquez, Meyers, Nivre, Pad{\'o}, {\v{S}}t{\v{e}}p{\'a}nek,
  Stra{\v{n}}{\'a}k, Surdeanu, Xue, and Zhang}]{conll2009}
Jan Haji{\v{c}}, Massimiliano Ciaramita, Richard Johansson, Daisuke Kawahara,
  Maria~Ant{\`o}nia Mart{\'\i}, Llu{\'\i}s M{\`a}rquez, Adam Meyers, Joakim
  Nivre, Sebastian Pad{\'o}, Jan {\v{S}}t{\v{e}}p{\'a}nek, Pavel
  Stra{\v{n}}{\'a}k, Mihai Surdeanu, Nianwen Xue, and Yi~Zhang. 2009.
\newblock The {C}o{NLL}-2009 shared task: Syntactic and semantic dependencies
  in multiple languages.
\newblock In \emph{Proceedings of the CoNLL}, pages 1--18.

\bibitem[{Hao et~al.(2019)Hao, Dong, Wei, and Xu}]{hao-etal-2019-visualizing}
Yaru Hao, Li~Dong, Furu Wei, and Ke~Xu. 2019.
\newblock Visualizing and understanding the effectiveness of {BERT}.
\newblock In \emph{Proceedings of the EMNLP}, pages 4141--4150.

\bibitem[{Hendrickx et~al.(2010)Hendrickx, Kim, Kozareva, Nakov,
  {\'O}~S{\'e}aghdha, Pad{\'o}, Pennacchiotti, Romano, and
  Szpakowicz}]{hendrickx2009semeval}
Iris Hendrickx, Su~Nam Kim, Zornitsa Kozareva, Preslav Nakov, Diarmuid
  {\'O}~S{\'e}aghdha, Sebastian Pad{\'o}, Marco Pennacchiotti, Lorenza Romano,
  and Stan Szpakowicz. 2010.
\newblock {S}em{E}val-2010 task 8: Multi-way classification of semantic
  relations between pairs of nominals.
\newblock In \emph{Proceedings of the 5th International Workshop on Semantic
  Evaluation}, pages 33--38.

\bibitem[{Howard and Ruder(2018)}]{howard2018universal}
Jeremy Howard and Sebastian Ruder. 2018.
\newblock Universal language model fine-tuning for text classification.
\newblock \emph{arXiv preprint arXiv:1801.06146}.

\bibitem[{Jawahar et~al.(2019)Jawahar, Sagot, and Seddah}]{jawahar2019}
Ganesh Jawahar, Beno{\^\i}t Sagot, and Djam{\'e} Seddah. 2019.
\newblock What does {BERT} learn about the structure of language?
\newblock In \emph{Proceedings of the ACL}, pages 3651--3657.

\bibitem[{Kim et~al.(2019{\natexlab{a}})Kim, Dyer, and
  Rush}]{kim-etal-2019-compound}
Yoon Kim, Chris Dyer, and Alexander Rush. 2019{\natexlab{a}}.
\newblock Compound probabilistic context-free grammars for grammar induction.
\newblock In \emph{Proceedings of the ACL}, pages 2369--2385.

\bibitem[{Kim et~al.(2019{\natexlab{b}})Kim, Rush, Yu, Kuncoro, Dyer, and
  Melis}]{kim-etal-2019-unsupervised}
Yoon Kim, Alexander Rush, Lei Yu, Adhiguna Kuncoro, Chris Dyer, and G{\'a}bor
  Melis. 2019{\natexlab{b}}.
\newblock Unsupervised recurrent neural network grammars.
\newblock In \emph{Proceedings of the NAACL}, pages 1105--1117.

\bibitem[{Kitaev and Klein(2018)}]{kitaev-klein-2018-constituency}
Nikita Kitaev and Dan Klein. 2018.
\newblock Constituency parsing with a self-attentive encoder.
\newblock In \emph{Proceedings of the ACL}, pages 2676--2686.

\bibitem[{Kovaleva et~al.(2019)Kovaleva, Romanov, Rogers, and
  Rumshisky}]{Revealing19}
Olga Kovaleva, Alexey Romanov, Anna Rogers, and Anna Rumshisky. 2019.
\newblock Revealing the dark secrets of {BERT}.
\newblock \emph{CoRR}, abs/1908.08593.

\bibitem[{Kuncoro et~al.(2018)Kuncoro, Dyer, Hale, Yogatama, Clark, and
  Blunsom}]{kuncoro-etal-2018-lstms}
Adhiguna Kuncoro, Chris Dyer, John Hale, Dani Yogatama, Stephen Clark, and Phil
  Blunsom. 2018.
\newblock {LSTM}s can learn syntax-sensitive dependencies well, but modeling
  structure makes them better.
\newblock In \emph{Proceedings of the ACL}, pages 1426--1436.

\bibitem[{Luo et~al.(2019)Luo, Jiang, Belinkov, and
  Glass}]{luo-etal-2019-improving}
Hongyin Luo, Lan Jiang, Yonatan Belinkov, and James Glass. 2019.
\newblock Improving neural language models by segmenting, attending, and
  predicting the future.
\newblock In \emph{Proceedings of the ACL}, pages 1483--1493.

\bibitem[{Marcus et~al.(1993)Marcus, Santorini, and Marcinkiewicz}]{PTB93}
Mitchell~P. Marcus, Beatrice Santorini, and Mary~Ann Marcinkiewicz. 1993.
\newblock Building a large annotated corpus of english: The penn treebank.
\newblock \emph{Computational Linguistics}, 19(2):313--330.

\bibitem[{Merity et~al.(2017)Merity, Xiong, Bradbury, and Socher}]{MerityX0S17}
Stephen Merity, Caiming Xiong, James Bradbury, and Richard Socher. 2017.
\newblock Pointer sentinel mixture models.
\newblock In \emph{Proceedings of the ICLR}.

\bibitem[{Mikolov et~al.(2013)Mikolov, Sutskever, Chen, Corrado, and
  Dean}]{MikolovSCCD13}
Tomas Mikolov, Ilya Sutskever, Kai Chen, Gregory~S. Corrado, and Jeffrey Dean.
  2013.
\newblock Distributed representations of words and phrases and their
  compositionality.
\newblock In \emph{Proceedings of the NIPS}, pages 3111--3119.

\bibitem[{Pennington et~al.(2014)Pennington, Socher, and
  Manning}]{pennington-etal-2014-glove}
Jeffrey Pennington, Richard Socher, and Christopher Manning. 2014.
\newblock {G}love: Global vectors for word representation.
\newblock In \emph{Proceedings of the EMNLP}, pages 1532--1543.

\bibitem[{Peters et~al.(2018)Peters, Neumann, Iyyer, Gardner, Clark, Lee, and
  Zettlemoyer}]{PetersNIGCLZ18}
Matthew~E. Peters, Mark Neumann, Mohit Iyyer, Matt Gardner, Christopher Clark,
  Kenton Lee, and Luke Zettlemoyer. 2018.
\newblock Deep contextualized word representations.
\newblock In \emph{in Proceedings of the NAACL}, pages 2227--2237.

\bibitem[{Rabinovich et~al.(2017)Rabinovich, Stern, and
  Klein}]{rabinovich-etal-2017-abstract}
Maxim Rabinovich, Mitchell Stern, and Dan Klein. 2017.
\newblock Abstract syntax networks for code generation and semantic parsing.
\newblock In \emph{Proceedings of the ACL}, pages 1139--1149.

\bibitem[{Radford et~al.(2018)Radford, Narasimhan, Salimans, and
  Sutskever}]{GPT2018}
Alec Radford, Karthik Narasimhan, Tim Salimans, and Ilya Sutskever. 2018.
\newblock Improving language understanding by generative pre-training.
\newblock \emph{Technical Report}.

\bibitem[{Shen et~al.(2018{\natexlab{a}})Shen, Lin, Huang, and
  Courville}]{ShenLHC18}
Yikang Shen, Zhouhan Lin, Chin{-}Wei Huang, and Aaron~C. Courville.
  2018{\natexlab{a}}.
\newblock Neural language modeling by jointly learning syntax and lexicon.
\newblock In \emph{Proceedings of the ICLR}.

\bibitem[{Shen et~al.(2018{\natexlab{b}})Shen, Tan, Sordoni, and
  Courville}]{OrderedShenTSC18}
Yikang Shen, Shawn Tan, Alessandro Sordoni, and Aaron~C. Courville.
  2018{\natexlab{b}}.
\newblock Ordered neurons: Integrating tree structures into recurrent neural
  networks.
\newblock \emph{CoRR}, abs/1810.09536.

\bibitem[{Socher et~al.(2010)Socher, Manning, and
  Ng}]{Socher10learningcontinuous}
Richard Socher, Christopher~D. Manning, and Andrew~Y. Ng. 2010.
\newblock Learning continuous phrase representations and syntactic parsing with
  recursive neural networks.
\newblock In \emph{In Proceedings of the NIPS-2010 Deep Learning and
  Unsupervised Feature Learning Workshop}.

\bibitem[{Socher et~al.(2013)Socher, Perelygin, Wu, Chuang, Manning, Ng, and
  Potts}]{socher-etal-2013-recursive}
Richard Socher, Alex Perelygin, Jean Wu, Jason Chuang, Christopher~D. Manning,
  Andrew Ng, and Christopher Potts. 2013.
\newblock Recursive deep models for semantic compositionality over a sentiment
  treebank.
\newblock In \emph{Proceedings of the EMNLP}, pages 1631--1642.

\bibitem[{Tai et~al.(2015)Tai, Socher, and Manning}]{tai-etal-2015-improved}
Kai~Sheng Tai, Richard Socher, and Christopher~D. Manning. 2015.
\newblock Improved semantic representations from tree-structured long
  short-term memory networks.
\newblock In \emph{Proceedings of the ACL}, pages 1556--1566.

\bibitem[{Tenney et~al.(2019)Tenney, Xia, Chen, Wang, Poliak, McCoy, Kim,
  Durme, Bowman, Das, and Pavlick}]{Probing19}
Ian Tenney, Patrick Xia, Berlin Chen, Alex Wang, Adam Poliak, R.~Thomas McCoy,
  Najoung Kim, Benjamin~Van Durme, Samuel~R. Bowman, Dipanjan Das, and Ellie
  Pavlick. 2019.
\newblock What do you learn from context? probing for sentence structure in
  contextualized word representations.
\newblock In \emph{Proceedings of the ICLR}.

\bibitem[{Tjong Kim~Sang and De~Meulder(2003)}]{CoNLL2003}
Erik~F. Tjong Kim~Sang and Fien De~Meulder. 2003.
\newblock Introduction to the {C}o{NLL}-2003 shared task: Language-independent
  named entity recognition.
\newblock In \emph{Proceedings of the CoNLL}, pages 142--147.

\bibitem[{Vaswani et~al.(2017)Vaswani, Shazeer, Parmar, Uszkoreit, Jones,
  Gomez, Kaiser, and Polosukhin}]{vaswani2017attention}
Ashish Vaswani, Noam Shazeer, Niki Parmar, Jakob Uszkoreit, Llion Jones,
  Aidan~N Gomez, {\L}ukasz Kaiser, and Illia Polosukhin. 2017.
\newblock Attention is all you need.
\newblock In \emph{Advances in neural information processing systems}, pages
  5998--6008.

\bibitem[{Vig and Belinkov(2019)}]{Analyzing1906}
Jesse Vig and Yonatan Belinkov. 2019.
\newblock Analyzing the structure of attention in a transformer language model.
\newblock \emph{CoRR}, abs/1906.04284.

\bibitem[{Wang et~al.(2018)Wang, Pham, Yin, and Neubig}]{wang-etal-2018-tree}
Xinyi Wang, Hieu Pham, Pengcheng Yin, and Graham Neubig. 2018.
\newblock A tree-based decoder for neural machine translation.
\newblock In \emph{Proceedings of the EMNLP}, pages 4772--4777.

\bibitem[{Wang et~al.(2019)Wang, Lee, and Chen}]{wang-etal-2019-tree}
Yaushian Wang, Hung-Yi Lee, and Yun-Nung Chen. 2019.
\newblock Tree transformer: Integrating tree structures into self-attention.
\newblock In \emph{Proceedings of the EMNLP}, pages 1061--1070.

\bibitem[{Williams et~al.(2017)Williams, Drozdov, and Bowman}]{parse170901121}
Adina Williams, Andrew Drozdov, and Samuel~R. Bowman. 2017.
\newblock Learning to parse from a semantic objective: It works. is it syntax?
\newblock \emph{CoRR}, abs/1709.01121.

\bibitem[{Xia et~al.(2019)Xia, Li, Zhang, Zhang, Fu, Wang, and
  Si}]{XiaL0ZFWS19}
Qingrong Xia, Zhenghua Li, Min Zhang, Meishan Zhang, Guohong Fu, Rui Wang, and
  Luo Si. 2019.
\newblock Syntax-aware neural semantic role labeling.
\newblock In \emph{Proceedings of the AAAI}, pages 7305--7313.

\bibitem[{Yang et~al.(2019)Yang, Dai, Yang, Carbonell, Salakhutdinov, and
  Le}]{yang2019xlnet}
Zhilin Yang, Zihang Dai, Yiming Yang, Jaime Carbonell, Ruslan Salakhutdinov,
  and Quoc~V Le. 2019.
\newblock Xlnet: Generalized autoregressive pretraining for language
  understanding.
\newblock \emph{CoRR}, abs/1906.08237.

\bibitem[{Yazdani and Henderson(2015)}]{yazdani-henderson-2015-incremental}
Majid Yazdani and James Henderson. 2015.
\newblock Incremental recurrent neural network dependency parser with
  search-based discriminative training.
\newblock In \emph{Proceedings of the CoNLL}, pages 142--152.

\end{thebibliography}

\bibliographystyle{acl_natbib}

\end{document}